\documentclass{llncs} 

\usepackage[utf8]{inputenc} 
\usepackage[T1]{fontenc}    
\usepackage{graphicx}       
\usepackage{amsmath}        
\usepackage{amssymb}        
\usepackage{booktabs}       
\usepackage{url}            
\usepackage{hyperref}       
\usepackage{cite}           
\usepackage{multirow}       
\usepackage{tabularx}       
\usepackage{makecell}

\hypersetup{
    colorlinks=false,
    linkcolor=blue,
    filecolor=magenta,
    urlcolor=cyan,
    pdftitle={Restoring Rhythm: Punctuation Restoration Using Transformer Models for Bangla, a Low-Resource Language},
    }
\begin{document}

\title{Restoring Rhythm: Punctuation Restoration Using Transformer Models for Bangla, A Low-Resource Language}

\author{Md Obyedullahil Mamun\inst{1} \and
Md Adyelullahil Mamun\inst{2} \orcidID{0000-0002-6314-228X} \and
Arif Ahmad\inst{3} \orcidID{0000-0002-2649-8981} \and
Md. Imran Hossain Emu\inst{1}}
\authorrunning{M.O. Mamun et al.}

\institute{Bangladesh Army International University of Science and Technology (BAIUST), Cumilla, Bangladesh \\
\email{\{obyedullahilmamun,mdihemu\}@gmail.com}
\and
BRAC University, Dhaka, Bangladesh \\
\email{md.adyelullahil.mamun@g.bracu.ac.bd}
\and
North East University Bangladesh (NEUB), Sylhet, Bangladesh \\
\email{arif@neub.edu.bd}}

\maketitle

\begin{abstract}

Punctuation restoration enhances the readability of text and is critical for post-processing tasks in Automatic Speech Recognition (ASR), especially for low-resource languages like Bangla. In this study, we explore the application of transformer-based models, specifically XLM-RoBERTa-large, to automatically restore punctuation in unpunctuated Bangla text. We focus on predicting four punctuation marks-period, comma, question mark, and exclamation mark across diverse text domains. To address the scarcity of annotated resources, we constructed a large, varied training corpus and applied data augmentation techniques. Our best performing model, trained with an augmentation factor of $\alpha = 0.20$, achieves an accuracy of 97.1\% on the News test set, 91.2\% on the Reference set, and 90.2\% on the ASR set.
 Results show strong generalization to reference and ASR transcripts, demonstrating the model’s effectiveness in real-world, noisy scenarios. This work establishes a strong baseline for Bangla punctuation restoration and contributes publicly available datasets and code to support future research in low-resource NLP.

\keywords{punctuation restoration \and punctuation marks \and deep learning \and transformer models \and natural language processing.}
\end{abstract}

\section{Introduction}
Punctuation restoration is a critical post-processing step that enhances the readability and usability of ASR-generated transcripts, supporting a range of downstream Natural Language Processing (NLP) tasks such as translation, summarization, and sentiment analysis \cite{jones_measuring_2003} \cite{matusov_improving_2007}. Without proper punctuation, the semantic boundaries between sentences become blurred, leading to ambiguity and reduced effectiveness of NLP pipelines.

Early models addressed this challenge using lexical features and statistical methods, such as Conditional Random Fields (CRF) \cite{lu_better_nodate} \cite{zhang_punctuation_nodate}, trained on large-scale corpora. The evolution of deep learning introduced more effective techniques like Long Short-Term Memory (LSTM) networks, Convolutional Neural Networks (CNNs), and, more recently, transformer-based models \cite{che_punctuation_nodate} \cite{gale_experiments_2017} \cite{zelasko_punctuation_2018} \cite{wang_self-attention_2018}.

Despite the proven success of transformer models like BERT \cite{devlin_bert_2019} and RoBERTa \cite{liu_roberta_2019} in various NLP tasks, their application to punctuation restoration in low-resource languages-such as Bangla-remains limited. The main challenges include the scarcity of annotated corpora, lack of standardized benchmarks, and domain mismatch between training (clean, structured text) and real-world use cases (noisy, ASR outputs).

To address these issues, we constructed a large, diverse Bangla dataset using publicly available sources, including leading newspapers such as Prothom Alo and The Daily Star, along with book transcripts and online platforms. Additionally, we established strong baselines for punctuation restoration across different text types, including structured news articles, general reference texts, and noisy ASR transcriptions. 
The primary contributions of this research are as follows: 
\begin{enumerate}
\item Exploring the use of transformer-based language models for punctuation restoration in Bangla.
\item Proposing a novel strategy for improving model performance using data augmentation techniques.
\item Creating and evaluating training datasets for Bangla and providing benchmark results for this task.
\item Addressing the restoration of four key punctuation marks: period ($\vert$), comma (,), question mark (?), and exclamation mark (!).
\item Making the source code and datasets publicly available\footnotemark{} to facilitate future research.
\end{enumerate}

\footnotetext{\url{https://github.com/Obyedullahilmamun/Punctuation-Restoration-Bangla}}

\section{Literature Review}
Punctuation restoration has significantly advanced with the introduction of transformer-based architectures, which consistently outperform traditional CNN and RNN models in accuracy and robustness. For instance, \cite{wang_self-attention_2018} highlights how transformers improved joint punctuation prediction and speech-to-text translation, significantly boosting transcription quality.

Transformers are particularly effective for sequence labeling tasks due to their self-attention mechanisms, which capture long-range dependencies and contextual cues \cite{vaswani_attention_2023}. Their ability to process input in parallel further improves training efficiency \cite{kim_fastformers_2020}. For punctuation restoration, these characteristics are crucial, especially when punctuation cues appear far from their relevant positions in the sentence \cite{bahdanau_neural_2016}.

However, despite these advantages, Bangla punctuation restoration remains relatively underexplored \cite{bijoy_advancing_2023}. Most Bangla studies rely on monolingual architectures or limited resources. The lack of large, annotated corpora and standardized evaluation metrics poses significant challenges for training and benchmarking models.

Recent research has proposed using data augmentation to mitigate data scarcity. Inspired by general strategies \cite{wei_eda_2019}, techniques like synonym replacement, random insertion, and back-translation \cite{sennrich_improving_2016} have been adapted for sequence tasks. Nonetheless, Bangla’s rich morphology makes naive augmentation methods problematic-e.g., modifying suffixes or introducing incoherent phrases. Linguistically informed methods that preserve syntax and morphology are thus more effective \cite{tariquzzaman_bda_2024}. Additionally, techniques originally designed for text classification tasks \cite{fadaee_data_2017} have been successfully adapted and refined for sequence labeling, leading to notable improvements in model robustness. These strategies, combined with multilingual modeling, contribute to the development of more generalizable systems for Bangla punctuation restoration.

Multilingual transformer models like XLM-RoBERTa have shown promise in low-resource settings \cite{vaswani_attention_2023}. These models leverage shared embeddings across languages and can transfer knowledge from high-resource to low-resource languages. Additionally, fine-tuning techniques such as adapter layers and parameter-efficient methods \cite{howard_universal_nodate} further enhance their applicability in resource-constrained scenarios.

Moreover, the literature stresses the importance of careful preprocessing tokenization, normalization, noise reduction-and rigorous evaluation using precision, recall, and F1-score \cite{bahdanau_neural_2016}\cite{powers_evaluation_2020}. These practices contribute to the development of robust models capable of generalizing across diverse textual domains.

In conclusion, the literature supports the growing relevance of transformer-based architectures, multilingual modeling, and intelligent data augmentation for punctuation restoration in low-resource languages like Bangla. Our work builds upon these strategies to create a strong foundation for future research.
\section{Datasets}
In order to advance punctuation restoration for Bangla, particularly given its status as a low-resource language, we developed a comprehensive dataset composed of diverse textual sources. The primary corpus was derived from a publicly available dataset of Bangla newspaper articles \cite{khatun_subword_2019}, which provided the bulk of the training material. Specifically, from a total of approximately 2.1 million tokens, roughly 1.3 million tokens were sourced from this reference corpus. This substantial component of the dataset ensured a foundational linguistic variety representative of contemporary written Bangla.

To broaden the range of punctuation phenomena, especially the occurrence of exclamation marks, we incorporated additional texts from literary and expressive domains. These sources, which frequently feature more emotive and dynamic punctuation usage, included narrative and literary websites such as \textit{Kishor Alo} (www.kishoralo.com), \textit{Kali O Kalam} (www.kaliokalam.com), and \textit{E Banglalibrary} (www.ebanglalibrary.com), among others. 

\begin{table}[htbp]
\centering
\caption{Bangla Dataset Distribution Summary (Token Counts)}
\label{tab:dataset_distribution}
\begin{tabular}{@{}lrrrrrr@{}}
\toprule
\textbf{Dataset} & \textbf{Total} & \textbf{Period} & \textbf{Comma} & \textbf{Question} & \textbf{Exclamation} & \textbf{Other (O)} \\
\midrule
Train & 2177058 & 162884 & 103824 & 9745 & 4896 & 1895709 \\
Dev & 207313 & 15885 & 8944 & 794 & 404 & 181286 \\
Test (News) & 104373 & 7369 & 5004 & 387 & 315 & 91298 \\
Test (Ref.) & 12669 & 1374 & 666 & 217 & 186 & 10226 \\
Test (ASR) & 10929 & 1150 & 561 & 178 & 151 & 8889 \\
\bottomrule
\end{tabular}
\end{table}

These supplemental materials enriched the dataset with instances of more nuanced punctuation patterns, thereby creating a more balanced and challenging test bed for punctuation restoration. 

\textbf{Dataset Structure and Annotation:}
The dataset was partitioned into training, development, and testing splits, as summarized in Table \ref{tab:dataset_distribution}. The training set consists of approximately 2.17 million tokens, the development set includes around 207,000 tokens, and the combined testing sets amount to roughly 127,000 tokens. Manual punctuation annotation was conducted across all subsets, ensuring a comprehensive training resource that encapsulates a variety of linguistic forms and styles.

\textbf{Supplementary Datasets:}  
To evaluate model performance in spoken language scenarios, we integrated two additional datasets. First, Manual Transcriptions were derived from approximately 65 minutes of monologue-style Bangla short story readings. These texts, primarily sourced from domains where exclamation usage is more frequent, were manually annotated with punctuation to provide a benchmark for clean, human-produced transcriptions.

Second, ASR Transcriptions were obtained by processing the same 65 minutes of audio through Google Cloud's Speech API. Since the ASR output lacked punctuation, it was manually annotated to assess the model's capacity to handle noisy, automatically generated transcripts. This approach simulates real-world scenarios where models must restore punctuation in speech recognition outputs and other noisy text streams.

\subsection{Punctuation Marks}  
Our annotation framework focused on restoring five punctuation categories: Period ($\vert$), Comma (,), Question (?), Exclamation (!), and O (denoting the absence of punctuation). In both the manually created and ASR-generated datasets, these punctuation marks were comprehensively annotated to ensure consistency and enable robust model evaluation across text modalities.

Notably, while exclamation marks accounted for less than 1\% of tokens in the training and development sets, their proportion exceeded 1\% in certain test sets, particularly the Test (Ref.) and Test (ASR) subsets, due to the inclusion of literary and narrative sources where exclamations are more prevalent. In contrast, the news-based test set maintained a lower frequency of exclamation usage, reflecting the more formal and objective style of journalistic text.

\subsection{Data Preprocessing}  
Prior to model training, all textual data underwent preprocessing to ensure quality and consistency. This pipeline involved noise removal, formatting normalization, and subword tokenization using techniques such as Byte-Pair Encoding (BPE) implemented via the \texttt{subword-nmt} library. Given the complexity and low-resource nature of Bangla, subword tokenization was instrumental in handling the language's extensive vocabulary and script variants \cite{khatun_subword_2019}.

Custom Python scripts then converted all punctuation symbols into corresponding labels suitable for supervised learning tasks. Combined with the aforementioned tokenization and preprocessing steps, these measures provided a clean, standardized input that supported robust and reproducible model development.

\section{Methodology}
\begin{figure}[htbp] 
\centering
\includegraphics[width=.8\textwidth]{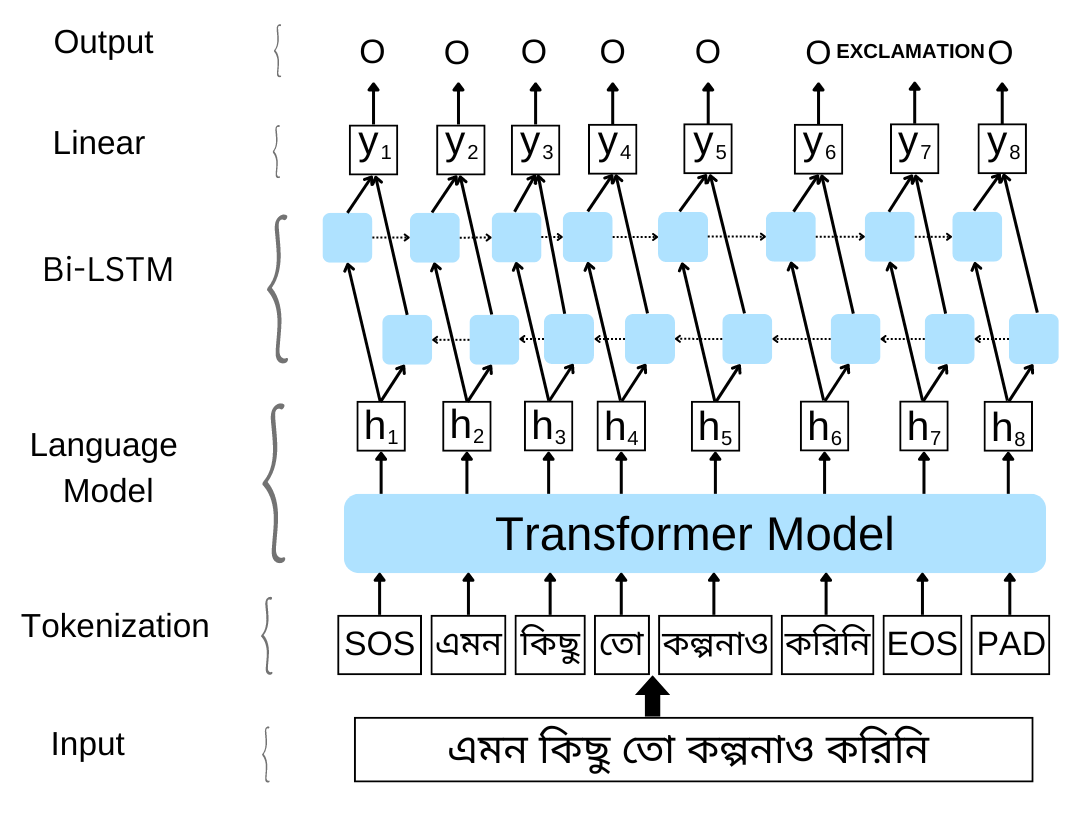}
\caption{Model architecture for punctuation restoration in Bangla}
\label{fig:model_architecture}
\end{figure}

Figure~\ref{fig:model_architecture} shows our XLM-RoBERTa-based architecture, selected for its demonstrated effectiveness in low-resource language processing. XLM-RoBERTa is pre-trained using a multilingual masked language model (MLM) objective, building on the structure of XLM but trained on a significantly more extensive and diverse dataset. This dataset includes over two terabytes of filtered Common Crawl data across one hundred languages \cite{wenzek_ccnet_2019}. XLM-RoBERTa's multilingual capabilities and pretraining on a diverse dataset \cite{conneau_unsupervised_2020} make it particularly well-suited for capturing the intricacies of Bangla text and punctuation. The model was further fine-tuned on the Bangla dataset to enhance its performance in punctuation restoration.

In the model training process, we selected appropriate hyperparameters, including the learning rate, batch size, and the number of training epochs. The methodology for using transformer models for punctuation restoration follows a similar approach as described in \cite{alam_punctuation_2020}.

In our approach, each token in the text is represented by a $d$-dimensional embedding vector obtained from the pre-trained language model. This embedding is passed into a Bidirectional Long Short-Term Memory (BiLSTM) layer with $h$ hidden units. The BiLSTM architecture enables the model to utilize both past and future context when making predictions, which is crucial for accurately restoring punctuation. The outputs from the forward and backward passes of the LSTM layers are concatenated at each time step. These concatenated outputs are then processed by a fully connected layer with five output neurons. Each neuron corresponds to one of the four punctuation marks (period, comma, question mark, exclamation mark) and an additional $O$ token representing non-punctuation.

We propose a data augmentation method inspired by \cite{wei_eda_2019} and customized for our dataset. This method addresses common errors in Automatic Speech Recognition (ASR), such as Substitution, Insertion, and Deletion. Notably, our dataset had fewer instances of exclamation marks compared to other punctuation marks like commas, periods, and question marks. To balance this, we applied augmentation techniques to introduce more exclamations, although the distribution was not perfectly balanced.

We employed the following three augmentation techniques:
\begin{enumerate}
\item {Substitution}: Randomly replacing tokens with a special unknown token to simulate substitution errors.
\item {Deletion}: Randomly removing tokens to simulate ASR deletion errors.
\item {Insertion}: Inserting unknown tokens at random positions to mimic insertion errors.
\end{enumerate}

We hypothesize that the frequency of substitution, insertion, and deletion errors varies and impacts the model's performance in different ways. To manage this variability, we utilized three tunable parameters:
\begin{enumerate}
\item {Token Change Probability}: The overall probability of making a change to a token.
\item {Substitution Probability}: The probability of replacing a token with an unknown token.
\item {Deletion Probability}: The probability of removing a token from the sequence.
\end{enumerate}

The Insertion Probability was calculated as the remainder after accounting for the substitution and deletion probabilities. This approach allows for the controlled introduction of errors into the dataset, simulating real-world ASR imperfections and helping to improve the robustness of our model.

\section{Experiments}
Our punctuation restoration experiments employed pre-trained transformer-based models sourced from the HuggingFace Transformers library \cite{wolf_huggingfaces_2020}. Input text was pre-processed using model-specific tokenizers with Byte-Pair Encoding (BPE) \cite{sennrich_neural_2016}, enabling efficient representation of subwords and rare tokens in Bangla. Each input sequence was truncated or padded to a maximum length of 256 tokens, framed by special start- and end-of-sequence tokens. Padding tokens were masked during the attention mechanism to ensure the model focused on meaningful input.

The training procedure involved a mini-batch size of 8, with data shuffled prior to each epoch to mitigate overfitting and enhance model generalization. Learning rates were set to 5e-6 for large transformer models and 1e-5 for base models, as initial exploratory runs suggested stable convergence at these rates. The Adam optimizer \cite{kingma_adam_2017} was employed for a total of ten epochs, and the LSTM hidden dimension used for sequence modeling tasks was matched to the token embedding size to maintain architectural consistency. Unless otherwise specified, hyperparameters were retained at their default settings proposed by Devlin et al. \cite{devlin_bert_2019}. Model selection was performed by evaluating performance on the development set after each epoch, with the best-performing model advanced for final testing.

\subsection{Data Augmentation}  
To enhance the robustness and variability of the training data, we employed data augmentation strategies inspired by previously established methods. Augmentation strength was controlled by the parameter $\alpha$, which determines the proportion of tokens in a sentence eligible for augmentation. We experimented with three values: $\alpha = 0.10$, $\alpha = 0.15$, and $\alpha = 0.20$. For each setting, substitution and deletion operations were applied at fixed rates of 0.4.

These controlled perturbations introduced lexical and structural variations into the training data, effectively expanding the training distribution. This, in turn, enabled the model to generalize better to diverse and unseen inputs.

\subsection{Evaluation Datasets}  
The finalized model was evaluated on three test datasets, each reflecting distinct text domains and characteristics:\\
1. News: A curated collection of Bangla news articles representing structured, formal text.\\
2. Reference (Ref): A broad selection of reference texts capturing more general, multi-genre written Bangla.\\
3. ASR: Transcriptions produced by Automatic Speech Recognition (ASR) systems, offering insight into the model’s capacity to restore punctuation under noisy and less-controlled input conditions.\\

\section{Results and Discussion}
We evaluated the model's ability to restore four types of punctuation marks- comma, period, question, and exclamation across three distinct test datasets: News, Reference (Ref.), and ASR. Performance was measured using precision (P), recall (R), and F1-score (F1) for each punctuation category, as well as overall accuracy. Table \ref{tab:detailed_performance} presents a detailed summary of these results.

\begin{table*}[htbp]
\centering
\caption{Performance of XLM-RoBERTa-large on Bangla Punctuation Restoration Across Different Datasets with and without Augmentation}
\label{tab:detailed_performance}
\resizebox{\textwidth}{!}{%
\begin{tabular}{p{3.2cm} l ccc ccc ccc ccc ccc}  

\toprule
\textbf{Model} & \textbf{Test} & \multicolumn{3}{c}{\textbf{Comma}} & \multicolumn{3}{c}{\textbf{Period}} & \multicolumn{3}{c}{\textbf{Question}} & \multicolumn{3}{c}{\textbf{Exclamation}} & \multicolumn{3}{c}{\textbf{Overall}} \\
\cmidrule(lr){3-5} \cmidrule(lr){6-8} \cmidrule(lr){9-11} \cmidrule(lr){12-14} \cmidrule(lr){15-17}
& & P & R & F1 & P & R & F1 & P & R & F1 & P & R & F1 & P & R & F1 \\
\midrule

\multirow{3}{*}{\makecell{XLM-RoBERTa-large}} 
& News & 83.7 & 77.6 & 80.5 & 86.5 & 92.2 & 89.3 & 71.2 & 83.9 & 77.1 & 67.9 & 33.6 & 45.0 & 84.8 & 84.9 & 84.9 \\
& Ref.  & 55.9 & 47.4 & 51.3 & 74.2 & 79.4 & 76.7 & 53.6 & 74.1 & 62.2 & 51.2 & 32.7 & 40.0 & 66.4 & 66.7 & 66.5 \\
& ASR   & 54.9 & 49.7 & 52.1 & 69.5 & 75.0 & 72.1 & 47.7 & 71.3 & 57.2 & 52.2 & 37.7 & 43.8 & 62.4 & 65.0 & 63.6 \\

\midrule
\multirow{3}{*}{\makecell{XLM-RoBERTa-large \\ + Aug. ($\alpha=0.10$)}} 
& News & 82.7 & 77.2 & 79.9 & 86.5 & 91.6 & 89.0 & 77.7 & 81.1 & 79.3 & 64.4 & 36.1 & 46.3 & 84.6 & 84.5 & 84.5 \\
& Ref.  & 55.5 & 47.1 & 51.0 & 75.3 & 79.1 & 77.2 & 61.4 & 64.5 & 62.9 & 47.3 & 38.1 & 42.2 & 67.6 & 66.0 & 66.8 \\
& ASR   & 53.8 & 49.9 & 51.8 & 69.9 & 73.5 & 71.6 & 53.4 & 60.6 & 56.8 & 42.1 & 39.0 & 40.5 & 62.4 & 63.3 & 62.8 \\

\midrule
\multirow{3}{*}{\makecell{XLM-RoBERTa-large \\ + Aug. ($\alpha=0.15$)}} 
& News & 83.7 & 76.4 & 79.9 & 86.1 & 91.1 & 88.5 & 69.3 & 84.7 & 76.2 & 70.1 & 23.1 & 34.8 & 84.5 & 83.7 & 84.1 \\
& Ref.  & 56.6 & 46.8 & 51.2 & 74.3 & 77.5 & 75.9 & 52.7 & 69.5 & 60.0 & 58.5 & 22.0 & 32.0 & 67.0 & 64.2 & 65.6 \\
& ASR   & 55.3 & 49.0 & 51.9 & 70.2 & 73.0 & 71.6 & 45.8 & 70.7 & 55.6 & 59.4 & 27.1 & 37.2 & 62.9 & 62.8 & 62.9 \\

\midrule
\multirow{3}{*}{\makecell{XLM-RoBERTa-large \\ + Aug. ($\alpha=0.20$)}} 
& News & 83.3 & 75.8 & 79.4 & 86.5 & 91.0 & 88.7 & 75.8 & 81.1 & 78.4 & 64.2 & 27.9 & 38.9 & 84.8 & 83.4 & 84.1 \\
& Ref.  & 58.1 & 47.5 & 52.3 & 74.0 & 78.0 & 75.9 & 55.8 & 68.2 & 61.4 & 58.6 & 29.0 & 38.8 & 67.6 & 65.1 & 66.4 \\
& ASR   & 57.1 & 48.3 & 52.3 & 69.2 & 73.5 & 71.3 & 49.7 & 65.1 & 56.4 & 55.1 & 35.7 & 43.3 & 63.5 & 63.0 & 63.3 \\

\bottomrule
\end{tabular}%
}
\end{table*}



The model exhibited its strongest performance on the News dataset, likely reflecting the structured and formal nature of news text. In contrast, performance declined on both the Reference and ASR datasets, which contain more diverse language usage and greater variability in style and domain complexity.

A consistent challenge across all datasets was the detection of exclamation marks. This difficulty was evident not only in the Reference and ASR sets, which include more expressive and speech-derived content, but also in the News dataset. The primary cause appears to be the relatively low frequency of exclamation marks in the training data. With fewer examples available, the model struggled to learn robust patterns for accurately predicting this punctuation type.

\section{Ablation Study}

To assess the impact of data augmentation on model performance, we conducted an ablation study comparing the base model (XLM-RoBERTa-large without augmentation) against variants trained with varying augmentation strengths. The augmentation parameter $\alpha$ specifies the proportion of tokens in each sentence eligible for modification. We evaluated three augmentation levels: $\alpha = 0.10$, $\alpha = 0.15$, and $\alpha = 0.20$. In all cases, substitution and deletion operations were applied at fixed rates of 0.4.

Table~\ref{tab:detailed_performance} presents precision (P), recall (R), and F1-score (F1) across four punctuation categories-comma, period, question mark, and exclamation mark-evaluated on three test domains: News, Reference (Ref.), and ASR. This comparison allows us to isolate the effect of augmentation relative to the base model. The key observations from this ablation study are as follows:

\begin{enumerate}
    \item {News Test Set:} The base model performed strongly on clean news text, achieving an F1-score of 84.9\%. Augmented models with $\alpha = 0.10$, $\alpha = 0.15$, and $\alpha = 0.20$ achieved comparable performance (F1 = 84.5\%, 84.1\%, and 84.1\%, respectively), indicating that augmentation did not degrade performance in structured domains.
    
    \item {Reference Test Set:} On the more diverse Ref. dataset, augmentation led to measurable improvements. The overall F1-score improved from 66.5\% (no augmentation) to 66.8\% ($\alpha = 0.10$) and 66.4\% ($\alpha = 0.20$), with slight variation for $\alpha = 0.15$ (65.6\%), suggesting enhanced generalization to varied linguistic structures.
    
    \item {ASR Test Set:} Augmented models maintained or slightly improved performance on noisy ASR transcripts. The base model achieved 63.6\% F1, while models with $\alpha = 0.10$, 0.15, and 0.20 scored 62.8\%, 62.9\%, and 63.3\%, respectively. These results suggest stable performance under challenging conditions.
    
    \item {Low-Frequency Punctuation:} Augmentation particularly helped with rare punctuation marks. For example, the exclamation F1-score in the News set increased from 45.0\% (no augmentation) to 46.3\% ($\alpha = 0.10$). Similarly, for the Ref. and ASR sets, augmentation improved or stabilized F1-scores for question and exclamation marks.
\end{enumerate}

\subsection{Comparative Analysis of Test Results} 
The XLM-RoBERTa-large + Aug. ($\alpha = 0.20$) model demonstrated high accuracy on the News test set (97.1\%), but its performance declined on the Reference (91.2\%) and ASR (90.2\%) test sets. These results are summarized in Table~\ref{tab:accuracy}. This disparity suggests that the model performs well on conventional and well-edited text, but faces challenges when dealing with less structured, colloquial, or speech-derived content.

\begin{table}[htbp]
\centering
\caption{Model Accuracy on Test sets}
\label{tab:accuracy}
\begin{tabular}{@{}lc@{}}
\toprule
\textbf{Test} & \textbf{Accuracy (\%)} \\
\midrule
News & 97.1 \\
Ref. & 91.2 \\
ASR & 90.2 \\
\bottomrule
\end{tabular}
\end{table}

\subsection{Confusion Matrix and Error Analysis}  
\begin{figure*}[htbp]
\centering
\includegraphics[width=\textwidth]{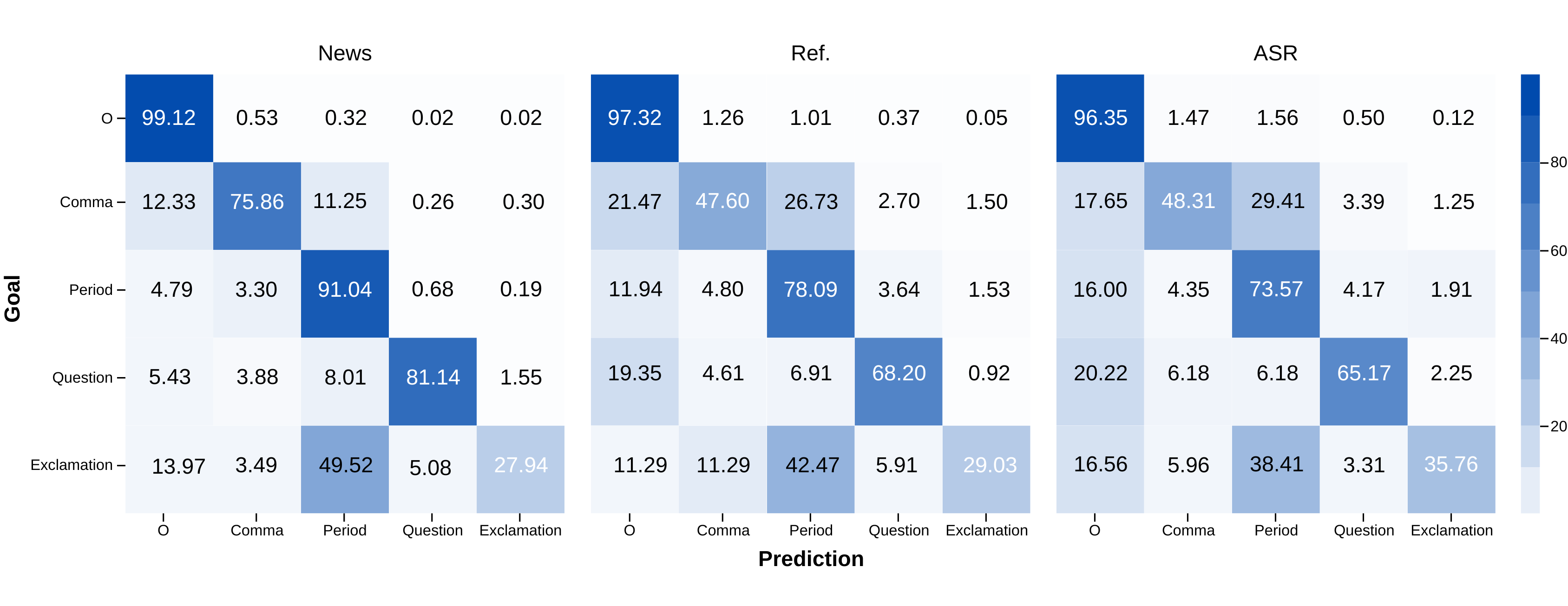}
\caption{Confusion matrices showing classification accuracy by punctuation type across test sets.}
\label{fig:confusion_matrix}
\end{figure*}

To gain deeper insights into the model’s misclassifications, we examined confusion matrices derived from each test dataset (Figure~\ref{fig:confusion_matrix}). These matrices reveal that the model consistently performs well in predicting the absence of punctuation-denoted as {O}-across all domains, which aligns with its high overall accuracy.

Despite overall satisfactory performance, the model exhibits notable difficulty in distinguishing between certain punctuation marks, particularly commas, periods, and question marks, within the Reference and ASR test sets. These challenges are reflected in increased off-diagonal activity in their respective confusion matrices, indicating a higher rate of misclassifications.

In the Reference set, for instance, only {47.60\%} of commas are correctly predicted, with frequent misclassifications as periods {26.73\%} and as the no-punctuation class {21.47\%}. Similarly, the ASR set shows a correct comma classification rate of {48.31\%}, with substantial confusion with periods {29.41\%} and the no-punctuation class {17.65\%}. Periods and question marks are also frequently confused. In the ASR data, question marks are correctly identified only {65.17\%} of the time, with significant misclassifications as no-punctuation {20.22\%} and commas {6.18\%}. Exclamation marks are particularly problematic, achieving only {35.76\%} accuracy, with high confusion with periods {38.41\%} and no-punctuation instances {16.56\%}.

This elevated confusion can be attributed to the inherent characteristics of the datasets. The ASR transcripts often contain disfluencies, inconsistent sentence boundaries, and prosodic ambiguities typical of spoken language, all of which complicate punctuation prediction. The Reference set, drawn from heterogeneous sources, exhibits diverse syntactic and stylistic conventions, further increasing ambiguity. In contrast, the News dataset displays a much cleaner pattern, with strong diagonal alignment in the confusion matrix-e.g., {91.04\%} accuracy for periods and {81.14\%} for question marks-highlighting the model's improved performance in formal, structured text.

To better optimize punctuation restoration for real-world ASR data, future work can explore targeted fine-tuning using speech-derived corpora annotated with both punctuation and disfluency markers. This would enable the model to learn contextual cues associated with spoken discourse. Domain-adaptive pretraining or fine-tuning on speech-style text-such as conversational transcripts, podcasts, or manually cleaned ASR outputs-can further improve robustness. Additionally, curriculum learning strategies, where the model is gradually exposed to increasingly noisy data, may enhance generalization to real-world conditions. Finally, integrating prosodic features aligned with audio (e.g., pause duration, pitch shifts) in a multimodal framework holds promise, though it remains beyond the scope of this work.

\section{Conclusion}
Our work has presented an effective approach to punctuation restoration for Bangla, an inherently low-resource language, by leveraging transformer-based architectures, specifically XLM-RoBERTa Large. Given the scarcity of annotated datasets and the limited body of prior work in this domain, our research aimed to address a critical gap in Bangla NLP. To that end, we introduced targeted data augmentation techniques designed to improve model robustness, particularly when dealing with noisy ASR outputs.

Our experimental results indicate that the proposed model excels in restoring punctuation accurately, even under challenging conditions such as speech-derived text. By providing a publicly available dataset and codebase, we seek to facilitate future research and foster collaboration within the NLP community. Through this contribution, we anticipate that the insights gained will not only enhance Bangla punctuation restoration capabilities but also serve as a blueprint for tackling similar challenges in other low-resource languages.


\bibliographystyle{splncs04}
\bibliography{paper.bib}

\end{document}